\documentclass[11pt]{article}

\usepackage[margin=1in]{geometry}
\usepackage[T1]{fontenc}
\usepackage{lmodern}
\usepackage{amsmath,amssymb}
\usepackage{booktabs}
\usepackage{array}
\usepackage{multirow}
\usepackage{graphicx}
\usepackage{xurl}
\usepackage[numbers,square,sort&compress]{natbib}
\usepackage[hidelinks]{hyperref}
\usepackage{caption}
\usepackage{setspace}
\newcommand{\Eref}{E_{\mathrm{ref}}}

\renewenvironment{abstract}{\section*{ABSTRACT}}{}
\newcommand{\alttext}[1]{\par\smallskip\noindent{\small Alt text: #1}}

\begin{document}

\title{Energy estimates of open-weight large language models across stated food weights: a dose-response evaluation}
\makeatletter
\begin{center}
{\LARGE \@title\par}
\vspace{1em}
{\large Donato Riccio, MSc\textsuperscript{1}, Fabrizio Maturo, PhD\textsuperscript{2}, Iulia Brezeanu, MSc\textsuperscript{3}\par}
\end{center}
\makeatother

\begin{singlespace}
{\small
\noindent\textsuperscript{1}Department of Engineering and Science, Universitas Mercatorum, Rome, Italy

\noindent\textsuperscript{2}Department of Economics, Statistics and Business, Universitas Mercatorum, Rome, Italy

\noindent\textsuperscript{3}Geometry Lab L.L.C-FZ, Dubai, United Arab Emirates

\medskip
\noindent Corresponding author: Donato Riccio, Department of Engineering and Science, Universitas Mercatorum, Piazza Mattei 10, 00186 Rome, Italy. Email: donato.riccio@studenti.unimercatorum.it

\medskip
\noindent\textbf{Keywords:} Large Language Models; Diet Records; Energy Intake; Portion Size; Artificial Intelligence

}
\end{singlespace}

\newpage
\begin{abstract}

\noindent\textbf{Objective:} To test whether energy estimates of open-weight large language models (LLMs) scale proportionally with an exactly stated food weight, and whether computing energy density times weight removes the error.

\noindent\textbf{Materials and Methods:} Under a prespecified protocol, seven open-weight LLMs estimated the energy of 168 randomly drawn foods from the USDA Food and Nutrient Database for Dietary Studies at nine stated weights, from 0.25 to 4 times the default amount. The primary outcome was the elasticity of the estimate with respect to weight (1: proportional scaling). A two-step estimate multiplied each model's own stated energy density by the weight in code.

\noindent\textbf{Results:} All seven models compressed their estimates, with elasticities from 0.548 to 0.979 (every 95\% CI below 1); four compressed severely (0.548 to 0.709). These four overestimated a quarter of the default amount by 47\% to 130\% and underestimated four times it by 20\% to 34\%. Supplying the density restored near-proportional scaling in three of the four, and the two-step estimate cut the typical error factor at these extremes from 1.47--2.43 to 1.20--1.38. Compression persisted at full precision and grew in five of seven models under a plain question.

\noindent\textbf{Discussion:} The models hold the energy densities needed for proportional scaling and answer near a typical serving, reproducing the flat-slope syndrome of dietary recall. Accuracy at usual portions does not reveal it.

\noindent\textbf{Conclusion:} Food-logging tools should compute the energy of weighed entries as density times weight, and evaluations should test fixed foods across stated weights.
\end{abstract}

\newpage
\section*{BACKGROUND AND SIGNIFICANCE}

People who track what they eat increasingly describe a meal in words and let a large language model (LLM) turn the description into energy and nutrients, and LLM assistants that analyze meals and give dietary guidance are being built and evaluated \cite{lu2026purrfessor}. Informatics work on free-text dietary data has classified patient-generated meal logs against nutritional goals \cite{hu2026meallogs} and extracted dietary-supplement information from clinical records with LLMs \cite{zhan2025ramie}. Evaluations have measured accuracy over meal descriptions with the amounts people report eating: on the NutriBench benchmark, the best model estimates the carbohydrate of 66.8\% of meal descriptions to within 7.5 g \cite{dhaliwal2025nutribench}. Many of the users who depend most on these numbers weigh their food. People with diabetes who count carbohydrates, people managing their weight and athletes type an exact weight, such as ``180 g of rice''. For them, an estimate is useful only if it scales with the weight: twice the food must give twice the energy.

Proportional scaling cannot be taken for granted. In dietary recall, people over-report small intakes and under-report large ones, a pattern described as the flat-slope syndrome \cite{gersovitz1978}. LLMs that estimate nutrients from food photographs show the same pattern: underestimation grows with portion size, with bias slopes of $-0.23$ to $-0.50$ \cite{fridolfsson2025}, and estimated meal weights cluster near a typical size \cite{ohara2025}. In these image studies the model must first judge the weight from pixels, so a compressed weight judgment and a compressed energy calculation cannot be told apart. Vision-language models also under-use a scale that is supplied to them: when every quantity in a metric physics question about a video is rescaled, their answers move only part of the way, with median log-log response slopes of 0.69 to 0.94 \cite{tan2026equisd}.

Benchmarks such as NutriBench score each meal at one amount and therefore cannot detect a scaling failure. An early version of NutriBench found a larger carbohydrate error per serving for multiples of a serving than for single servings, and its authors suggested that models trained on values per 100 g struggle to use them for other serving sizes \cite{hua2024nutribenchv1}; that comparison used an unsigned error, so it cannot show the direction of the error or whether estimates scale in proportion. A study of image-based estimation found that a Japanese-dietitian persona lowered estimated portions consistently, read the effect as a shift of the model's portion prior, and named gram-specified servings as a natural next experiment \cite{nakagawa2026}. How the energy estimates of text LLMs change across a range of exactly stated weights for the same foods has not been measured.

\section*{OBJECTIVE}

We tested whether open-weight LLM energy estimates scale in proportion to an exactly stated food weight, whether the models know the required energy densities, and whether computing density times weight removes the error. Because the stated weight is the only information about scale in the prompt, every departure from proportionality is an error of the model, and the elasticity of the estimate with respect to the weight measures it in one number that is unaffected by how well the model knows the food's energy density.

\section*{MATERIALS AND METHODS}

\subsection*{Study design}

Each food was presented at nine stated weights. The study had two stages. A development stage on 56 foods and four models fixed the prompts, the outcome, the analysis code and the decision thresholds (Supplementary File 1). The confirmatory stage reported here used 168 new foods and seven models. Its protocol, prompts, generation code and analysis code were fixed and recorded with SHA-256 hashes before any confirmatory response existed; the robustness arms and the token-probability analysis followed the same rule. Reporting follows the TRIPOD-LLM guideline (Supplementary File 2) \cite{gallifant2025tripodllm}.

\subsection*{Foods and reference energy}

Foods came from the US Department of Agriculture (USDA) Food and Nutrient Database for Dietary Studies (FNDDS) 2021--2023, distributed in the FoodData Central survey release of October 31, 2024 \cite{fndds2023,fdc2019}. FNDDS gives each food's energy per 100 g and a set of portion weights. One of them, the ``Quantity not specified'' portion, is the weight the survey assigns when a respondent reports no amount; we call it the default amount $q$. The sampling frame contained the FNDDS foods of seven What We Eat in America (WWEIA) groups of solid foods \cite{wweia2023}: cheese and yogurt, protein foods, mixed dishes, grains, snacks and sweets, fruit, and vegetables. Eligible foods had a positive energy value and a default amount of at least 20 g. We drew 24 foods per group uniformly at random (seed 20261004) after excluding the development foods, giving 168 foods. Prompts used the FNDDS description of each food with the survey qualifiers ``NFS'' and ``NS as to'' removed where they formed a separate part of the description. Energy densities ranged from 20 to 754 kcal per 100 g (median 191) and default amounts from 20 to 315 g (median 88 g). For a food with FNDDS energy density $D$ (kcal per 100 g), the reference energy of a stated weight of $g$ grams is $\Eref = gD/100$.

\subsection*{Prompts}

Each food generated 14 prompts in four conditions (Supplementary File 1 gives both system prompts verbatim). A system prompt cast the model as the nutrition estimator of a food-logging app and required a JSON object with the served weight, one item per food component with its weight and energy, and the assumptions made. The stated-weight condition asked about ``\textit{g} g of \textit{food}, weight as eaten.'' at nine weights spaced evenly on a logarithmic scale from $0.25q$ to $4q$: multipliers 0.25, 0.35, 0.5, 0.71, 1, 1.41, 2, 2.83 and 4, rounded to whole grams. This range holds 77\% of the other portion weights that FNDDS lists for the 168 foods (Supplementary File 1, section S8). The no-amount condition gave the food name alone, so the model chose its own serving, which we call the model's default serving. The density condition asked, under a second system prompt, for the energy density of the food in kilocalories per 100 g. The supplied-density condition repeated the stated-weight prompt at $0.25q$, $q$ and $4q$ followed by ``It provides \textit{D} kcal per 100 g.'', with the FNDDS value rounded to a whole number. The system prompt was written for an unpublished pilot and used unchanged; it was not tuned on the foods or outcomes of this study.

\subsection*{Models and generation}

Seven open-weight instruction-tuned models from six developers spanned 8 to 70 billion parameters, dense and mixture-of-experts architectures, and one reasoning model (Table~\ref{tab:models}) \cite{qwen35,gemma4,mistralsmall32,grattafiori2024llama3,llama33card,llama31card,gptoss2025,abdin2024phi4}. The development stage used Qwen3.5-35B-A3B, Gemma-4-26B-A4B, Mistral-Small-3.2-24B and Llama-3.1-8B; the confirmatory stage added Llama-3.3-70B, gpt-oss-20b and Phi-4 to span developers, sizes and reasoning. Each ran as a 4-bit conversion for MLX, Apple's machine-learning framework, at a pinned repository revision on two Mac Pro computers (Apple M2 Ultra, 128 GB of unified memory), using MLX 0.32.2 and mlx-lm 0.31.3 \cite{mlx2023}. Supplementary File 1 (Tables S4 and S5) lists the revisions, release dates and stated training-data dates. gpt-oss-20b ran in its native MXFP4 format at reasoning effort ``low'', the setting closest to a direct answer, and only its final answer channel was parsed; Qwen3.5-35B-A3B ran with thinking disabled. Sampling followed each model's published generation configuration; gpt-oss-20b and Phi-4 set none and ran at temperature 1.0. Each prompt received three sampled responses and one greedy response, with up to three attempts when a response did not parse. A response was valid when it parsed as the required JSON object with positive energy values; one repair was accepted, a closing brace missing at the end of an otherwise complete object. Generating the 9,408 confirmatory responses of a model took 9,422 to 9,519 calls and 1.0 to 10.9 hours (Supplementary File 1, Table S20). The energy estimate $\hat{E}$ of a response is the sum of the energies of its items.

\begin{table}[t]
\centering
\caption{Models. Parameters are total (active per token for mixture-of-experts models). Sampling gives temperature $T$, nucleus probability $p$ and top-$k$ ($k$ = 0: off). Revisions are in Supplementary File 1.}
\label{tab:models}
\small
\begin{singlespace}
\begin{tabular}{@{}llllr@{}}
\toprule
Model & Developer & Parameters & Architecture & Sampling ($T$, $p$, $k$) \\
\midrule
Qwen3.5-35B-A3B & Alibaba & 35B (3B) & mixture of experts & 1.0, 0.95, 20 \\
gpt-oss-20b & OpenAI & 21B (3.6B) & mixture of experts, reasoning & 1.0, 1.0, 0 \\
Gemma-4-26B-A4B & Google & 26B (3.8B) & mixture of experts & 1.0, 0.95, 64 \\
Mistral-Small-3.2-24B & Mistral AI & 24B & dense & 0.15, 1.0, 0 \\
Llama-3.3-70B & Meta & 70B & dense & 0.6, 0.9, 0 \\
Phi-4 & Microsoft & 14B & dense & 1.0, 1.0, 0 \\
Llama-3.1-8B & Meta & 8B & dense & 0.6, 0.9, 0 \\
\bottomrule
\end{tabular}
\end{singlespace}
\end{table}

\subsection*{Outcomes}

The energy ratio of an estimate is $r = \hat{E}/\Eref$. For food $f$ at multiplier $m$, let $y_{fm}$ be the mean of $\log r$ over the three sampled responses and $x_{fm} = \log(g_{fm}/q_f)$. The compression slope $b$ is the food-centred least-squares slope of $y$ on $x$ over the nine stated weights,
\begin{equation}
b = \frac{\sum_{f,m} (x_{fm} - \bar{x}_f)(y_{fm} - \bar{y}_f)}{\sum_{f,m} (x_{fm} - \bar{x}_f)^2},
\end{equation}
where $\bar{x}_f$ and $\bar{y}_f$ are means over the weights of food $f$. The elasticity $\beta = 1 + b$ is the proportional change in the estimate per proportional change in the stated weight; at $\beta = 0.7$, doubling the stated weight raises the estimate by 62\%. Food centring removes any error in a model's density for a food that stays constant across weights, so a model that holds a wrong density but scales it correctly has $\beta = 1$. We call a model compressing when the 95\% CI of $\beta$ lies below 1, and severely compressing when, in addition, $\beta \le 0.95$; at that threshold, fixed in the development stage, doubling the stated weight raises the estimate by 93\%. A compressing model's estimates vary less than the stated weight around a middle portion; because the curves cross the reference near the default amount $q$ (Table~\ref{tab:main}), we call this compression toward a typical serving.

The two-step estimate of a stated weight is $\hat{E}_2 = \hat{D} g / 100$, where $\hat{D}$ is the geometric mean of the model's three sampled answers to the density question for that food. It scales in exact proportion to the stated weight by construction, and its error is the model's density error. Accuracy is the absolute log error $|\log(\hat{E}/\Eref)|$ averaged over foods (mean absolute log error, MALE); $e^{\mathrm{MALE}}$ is the typical multiplicative error factor.

\subsection*{Confirmatory claims and criteria}

Four claims and their criteria were fixed before data collection.
\begin{enumerate}
\item \emph{Prevalence.} The 95\% CI of $\beta$ lies below 1 in at least five of the seven models.
\item \emph{Replication of severity.} Mistral-Small-3.2-24B and Llama-3.1-8B, the two models that compressed most in the development stage, compress severely on the new foods.
\item \emph{Supplied density.} In every severely compressing model, supplying the energy density raises the elasticity measured over $0.25q$, $q$ and $4q$: the paired 95\% CI of the difference lies above 0.
\item \emph{Two-step estimate.} In every severely compressing model, the two-step estimate has a lower MALE than the direct estimate at $0.25q$ and at $4q$: the paired 95\% CI of the difference (direct minus two-step) lies above 0 at both weights.
\end{enumerate}
A model needed a valid share of at least 95\% of its responses to enter the counts. Three secondary analyses were also fixed in advance: the elasticity difference between Llama-3.3-70B and Llama-3.1-8B, the shape of the dose-response curves, and the models' default servings. The comparison of food groups is exploratory.

\subsection*{Statistical analysis}

The food is the independent unit. Every CI is a percentile interval from 2,000 bootstrap resamples of foods (seed 20261004); paired comparisons resample foods jointly. Responses that stayed invalid after three attempts were excluded. The elasticity used every food-weight cell with at least one valid sampled response; 7 of the 10,584 cells had none (Supplementary File 1, section S5). For curve shape, we report the mean log-ratio curve with a 95\% simultaneous bootstrap band, functional principal components of the food-centred curves \cite{ramsay2005fda}, a quadratic term in centred log weight, and a permutation test of the curves across the seven food groups. For default servings, we compare the model's served weight in the no-amount condition with $q$ and test whether the dose-response curve pivots at the model's own default serving (Supplementary File 1, section S5). A post hoc descriptive check classified each supplied-density answer by its relation to the energy implied by the supplied density. Analyses used Python 3.12.13 with NumPy 2.5.3, pandas 3.0.6 and SciPy 1.18.1. The code was developed with the assistance of a generative AI tool (see Acknowledgments); before any confirmatory response existed, the analysis code recovered a compression slope of $-0.30$ planted in synthetic responses (estimate $-0.301$).

\subsection*{Robustness analyses}

Three robustness arms were fixed and hash-recorded before their responses existed. The full-precision arm repeated the 1,512 stated-weight prompts with the unquantized bfloat16 weights of Llama-3.1-8B, Phi-4 and Mistral-Small-3.2-24B; Llama-3.3-70B does not fit in memory at full precision. The plain-wording arm removed the system prompt and the answer format: all seven models answered ``How many kilocalories are in \textit{g} g of \textit{food}? Reply with the number only.'' for the same foods and weights. In both arms, compression replicated when $b$ was at most $-0.05$ with its 95\% CI below 0. The reasoning arm put the plain question to the four severely compressing models with the instruction ``Think step by step, then give your final answer on the last line in the form `Energy: <number> kcal'.'' at $0.25q$, $q$ and $4q$, with greedy decoding, and compared the three-weight elasticity with that of the greedy answers to the plain question.

\subsection*{Token probabilities}

Open weights expose the probability that a model assigns to every possible answer. For each prompt, we fixed the model's JSON answer up to the energy field of a single item named as in the prompt and enumerated the numbers the model could write next, token by token \cite{requeima2024llmp}, which gives a probability distribution over the energy values the model can write. For gpt-oss-20b, the reader conditioned on the model's greedy reasoning, which already states its answer. The reader was valid for a model when the elasticity of the distribution median lay within a prespecified tolerance of 0.05 of the elasticity of the sampled answers; this checks the aggregate scaling of the distributions, not each distribution. At the extreme weights ($0.25q$, $0.35q$, $2.83q$ and $4q$), we asked whether a model's probability splits between remembering and computing or sits on one compromise value. Two references came from the model's own distributions for the same food: the computed value $C$, its density answer times the stated weight, and the remembered value $R$, its energy for the food named alone. A switching account predicts a mixture of the two reference distributions; a compromise account predicts one value between them on the log scale. Each account has one weight, fitted per prompt, that is 1 for pure computing and 0 for pure remembering. The two accounts are two fitted families of distributions, and a model was classified as compromising when the 95\% CI of the mean difference in expected log-likelihood (compromise minus switching) lay above 0. The position $t = (\log v - \log R)/(\log C - \log R)$ of each candidate energy $v$ is 0 at $R$ and 1 at $C$. The distributions of the 56 development foods fixed these endpoints before any confirmatory distribution existed. Supplementary File 1 (section S7) gives the enumeration, smoothing, positive control and sensitivity analyses in full.

\subsection*{Ethical considerations}

The study used only publicly available data, a public-domain food composition database of the US Department of Agriculture, and the outputs of language models. It involved no human participants, animals or personal data and therefore did not require ethics review.

\section*{RESULTS}

\subsection*{Response quality}

Of 65,856 designed responses (2,352 prompts, four responses per prompt, seven models), every model returned a valid answer for at least 99.5\% (Table~\ref{tab:main}). The models restated the stated weight as the served weight to within 2\% in 98.6\% to 100\% of stated-weight responses, and in 94.6\% for gpt-oss-20b. Compression therefore arises after the models have read the weight correctly.

\subsection*{Compression across models}

\begin{figure}[t]
\centering
\includegraphics[width=\textwidth]{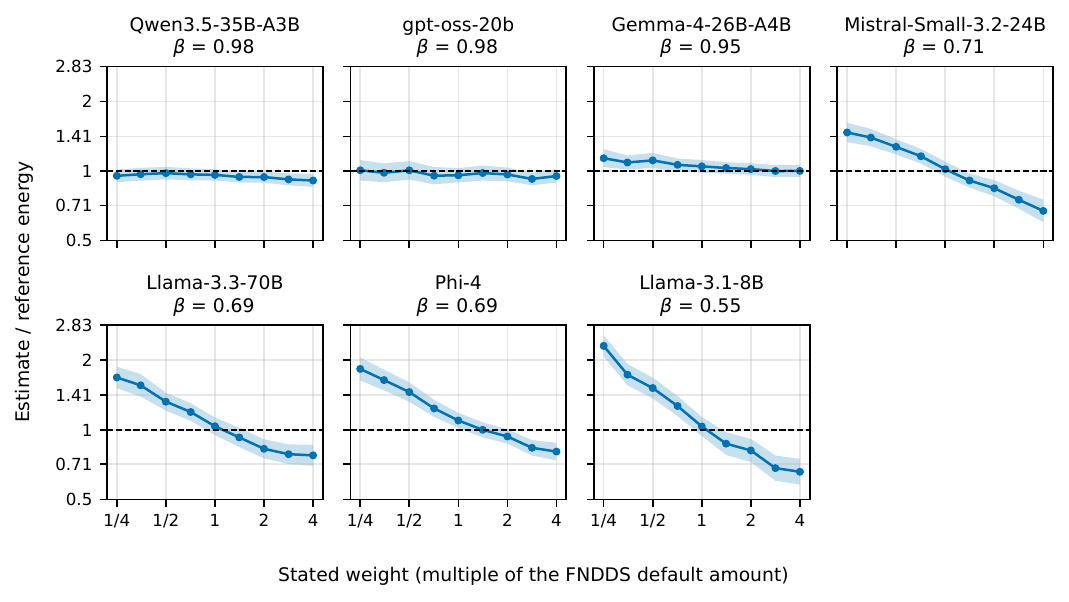}
\caption{Dose-response curves of energy estimates against stated weight on 168 foods. Each panel shows the geometric mean over foods of estimate / reference energy (points) at nine stated weights from a quarter to four times the FNDDS default amount, with a 95\% simultaneous food-bootstrap band (shaded). Both axes are logarithmic. Proportional scaling is the dashed horizontal line at 1; a curve that falls from left to right is compressed. Panels are ordered by elasticity. Curves use the foods with valid responses at all nine weights (167 for Mistral-Small-3.2-24B and 162 for Llama-3.3-70B; 168 otherwise).}
\alttext{Seven small panels, one per model, plotting the ratio of estimated to reference energy against stated weight on logarithmic axes. Three panels stay near the horizontal line at 1; four fall from 1.47 to 2.30 at the smallest weight to 0.66 to 0.80 at the largest.}
\label{fig:dose}
\end{figure}

All seven models compressed: the 95\% CI of the elasticity lay below 1 in seven of seven models, against a criterion of five (Table~\ref{tab:main}, Figure~\ref{fig:dose}), so Claim 1 is confirmed. Elasticities ranged from 0.548 to 0.979. Four models compressed severely: Mistral-Small-3.2-24B ($\beta = 0.709$), Llama-3.3-70B (0.694), Phi-4 (0.692) and Llama-3.1-8B (0.548). The two models named in advance are among them, so Claim 2 is confirmed. Qwen3.5-35B-A3B (0.979), gpt-oss-20b (0.978) and Gemma-4-26B-A4B (0.954) scaled close to proportionally with this prompt.

Severe compression is two-sided and centred near the default amount. At $q$, the four severely compressing models estimated 1.02 to 1.10 times the reference energy. At a quarter of the default amount they estimated 1.47 to 2.30 times it, and at four times the default amount 0.66 to 0.80 times it. For Mistral-Small-3.2-24B, doubling the stated weight raised the estimate by 63\%. Greedy decoding gave the same elasticities as sampling to within 0.022 in every model.

\begin{table}[t]
\centering
\caption{Scaling and density knowledge by model on 168 foods. $\beta$: elasticity of the energy estimate with respect to the stated weight (1: proportional), with 95\% CI from the food bootstrap. Ratio: geometric mean over foods of estimate / reference energy at a stated weight of $0.25q$, $q$ and $4q$. Greedy: elasticity of the greedy responses. Density: the model's answer to the density question against the FNDDS value, as the median ratio and the median absolute percentage error (MdAPE) over foods. Valid: share of the 9,408 designed responses that parsed. Density statistics of Gemma-4-26B-A4B use 166 foods, which gave a valid density answer.}
\label{tab:main}
\small
\begin{singlespace}
\begin{tabular}{@{}lccccccccc@{}}
\toprule
& & & \multicolumn{3}{c}{Ratio at} & & \multicolumn{2}{c}{Density} & \\
\cmidrule(lr){4-6}\cmidrule(lr){8-9}
Model & $\beta$ & 95\% CI & $0.25q$ & $q$ & $4q$ & Greedy & Ratio & MdAPE & Valid \\
\midrule
Qwen3.5-35B-A3B & 0.979 & 0.967--0.990 & 0.95 & 0.96 & 0.91 & 0.987 & 0.91 & 15.0\% & 100.0\% \\
gpt-oss-20b & 0.978 & 0.955--0.999 & 1.01 & 0.96 & 0.95 & 0.977 & 0.98 & 14.9\% & 100.0\% \\
Gemma-4-26B-A4B & 0.954 & 0.935--0.968 & 1.14 & 1.05 & 1.00 & 0.953 & 1.00 & 9.2\% & 99.9\% \\
Mistral-Small-3.2-24B & 0.709 & 0.678--0.741 & 1.47 & 1.02 & 0.67 & 0.722 & 1.00 & 12.1\% & 99.7\% \\
Llama-3.3-70B & 0.694 & 0.667--0.724 & 1.68 & 1.04 & 0.77 & 0.689 & 0.99 & 10.7\% & 99.5\% \\
Phi-4 & 0.692 & 0.664--0.722 & 1.83 & 1.10 & 0.80 & 0.714 & 1.00 & 13.2\% & 100.0\% \\
Llama-3.1-8B & 0.548 & 0.514--0.584 & 2.30 & 1.03 & 0.66 & 0.546 & 1.04 & 21.0\% & 99.8\% \\
\bottomrule
\end{tabular}
\end{singlespace}
\end{table}

\subsection*{Knowledge of energy density}

Asked directly, the models gave energy densities whose median ratio to the FNDDS value ranged from 0.91 to 1.04, with median absolute percentage errors of 9.2\% to 21.0\% (Table~\ref{tab:main}). Density knowledge did not separate the two groups of models. Mistral-Small-3.2-24B, Llama-3.3-70B and Phi-4 compressed severely with median absolute density errors of 10.7\% to 13.2\%, while Qwen3.5-35B-A3B and gpt-oss-20b scaled almost proportionally with errors of 15.0\% and 14.9\%. Llama-3.1-8B had both the largest density error (21.0\%) and the strongest compression. The severely compressing models therefore held the information needed for proportional scaling and left it unused.

\subsection*{Supplying the energy density}

Supplying the FNDDS density in the prompt raised the three-weight elasticity of every severely compressing model (Table~\ref{tab:fix}, Figure~\ref{fig:fix}A): from 0.718 to 0.973 for Mistral-Small-3.2-24B, from 0.720 to 0.995 for Llama-3.3-70B, and from 0.703 to 0.984 for Phi-4. The paired differences ranged from 0.227 to 0.281, and every 95\% CI lay above 0, so Claim 3 is confirmed. Llama-3.1-8B remained compressed with the density in front of it, rising from 0.548 to 0.776. In the post hoc check (Supplementary File 1, Table S8), six models returned an energy within 5\% of the value implied by the supplied density in 93.7\% to 97.0\% of answers. Llama-3.1-8B was correct in 23.1\% of answers; in 28.4\% it reported the supplied per-100 g figure itself as the energy of the stated weight, and in 15.6\% it was off by a factor of ten, so its estimates with the density supplied averaged 0.26 to 0.49 times the reference energy. Claim 3 concerns scaling, and for Llama-3.1-8B the higher elasticity came with much lower accuracy. Mistral-Small-3.2-24B, Llama-3.3-70B and Phi-4 therefore multiply correctly when the density is in the prompt and leave their own density knowledge unused when it is not.

\begin{table}[t]
\centering
\caption{Supplied density and two-step estimate in the four severely compressing models, 168 foods. Elasticity over the stated weights $0.25q$, $q$ and $4q$ without and with the FNDDS energy density in the prompt, and the paired difference. MALE: mean absolute log error of the direct estimate and of the two-step estimate (the model's own density answer times the stated weight) at $0.25q$ and $4q$, over the foods with a valid direct estimate at that weight (all 168, except 165 at $0.25q$ and 167 at $4q$ for Llama-3.3-70B). Intervals are 95\% CIs from the paired food bootstrap. For Llama-3.1-8B, the three-weight elasticity without the density coincides with its nine-weight elasticity (0.548).}
\label{tab:fix}
\small
\begin{singlespace}
\begin{tabular}{@{}lccccccc@{}}
\toprule
& \multicolumn{3}{c}{Elasticity ($0.25q$, $q$, $4q$)} & \multicolumn{2}{c}{MALE at $0.25q$} & \multicolumn{2}{c}{MALE at $4q$} \\
\cmidrule(lr){2-4}\cmidrule(lr){5-6}\cmidrule(lr){7-8}
Model & Without & With & Difference & Direct & Two-step & Direct & Two-step \\
\midrule
Mistral-Small-3.2-24B & 0.718 & 0.973 & 0.255 (0.218--0.292) & 0.467 & 0.204 & 0.518 & 0.204 \\
Llama-3.3-70B & 0.720 & 0.995 & 0.274 (0.244--0.305) & 0.570 & 0.185 & 0.433 & 0.183 \\
Phi-4 & 0.703 & 0.984 & 0.281 (0.251--0.312) & 0.663 & 0.246 & 0.386 & 0.246 \\
Llama-3.1-8B & 0.548 & 0.776 & 0.227 (0.144--0.303) & 0.887 & 0.319 & 0.580 & 0.319 \\
\bottomrule
\end{tabular}
\end{singlespace}
\end{table}

\subsection*{Two-step estimate}

\begin{figure}[t]
\centering
\includegraphics[width=\textwidth]{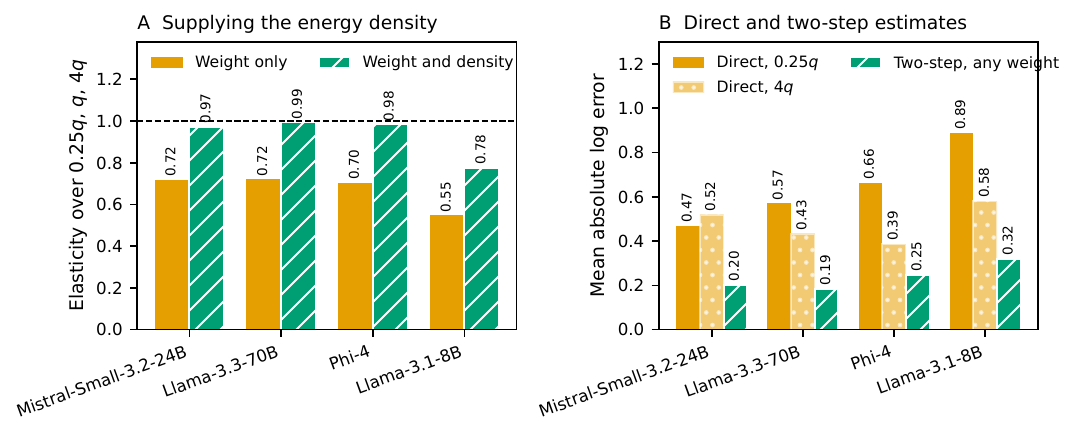}
\caption{Supplied density and two-step estimate in the four severely compressing models, 168 foods. (A) Elasticity of the energy estimate over the stated weights $0.25q$, $q$ and $4q$ with the weight only and with the FNDDS energy density added to the prompt; the dashed line at 1 is proportional scaling. (B) Mean absolute log error of the direct estimate at $0.25q$ and at $4q$, and of the two-step estimate, whose error over the same foods is the same at every stated weight. Table~\ref{tab:fix} gives the elasticity differences with 95\% CIs; the text gives the paired differences in mean absolute log error.}
\alttext{Two bar charts for four models. In panel A, elasticity rises from 0.703 to 0.720 without the density to 0.973 to 0.995 with it in three models, and from 0.55 to 0.78 in Llama-3.1-8B. In panel B, the two-step error bars are much shorter than the direct-estimate bars at both extreme weights in every model.}
\label{fig:fix}
\end{figure}

The two-step estimate was more accurate than the direct estimate at both extreme weights in every severely compressing model (Table~\ref{tab:fix}, Figure~\ref{fig:fix}B), so Claim 4 is confirmed. At a quarter of the default amount, the direct estimates had typical error factors of 1.60 to 2.43 and the two-step estimates 1.20 to 1.38; at four times the default amount, the direct estimates had error factors of 1.47 to 1.79. The paired differences in MALE ranged from 0.264 to 0.568 at $0.25q$ and from 0.140 to 0.315 at $4q$, with every 95\% CI above 0. The two values for Llama-3.3-70B (0.185 and 0.183) differ only because its paired foods number 165 at $0.25q$ and 167 at $4q$. In absolute terms (post hoc), the direct estimates of the severely compressing models missed the reference energy by a median of 18 to 45 kcal per food at $0.25q$ and 135 to 193 kcal at $4q$, against 5 to 6 kcal and 69 to 95 kcal in the near-proportional models (Supplementary File 1, Table S21).

\subsection*{Secondary analyses}

\begin{figure}[t]
\centering
\includegraphics[width=0.85\textwidth]{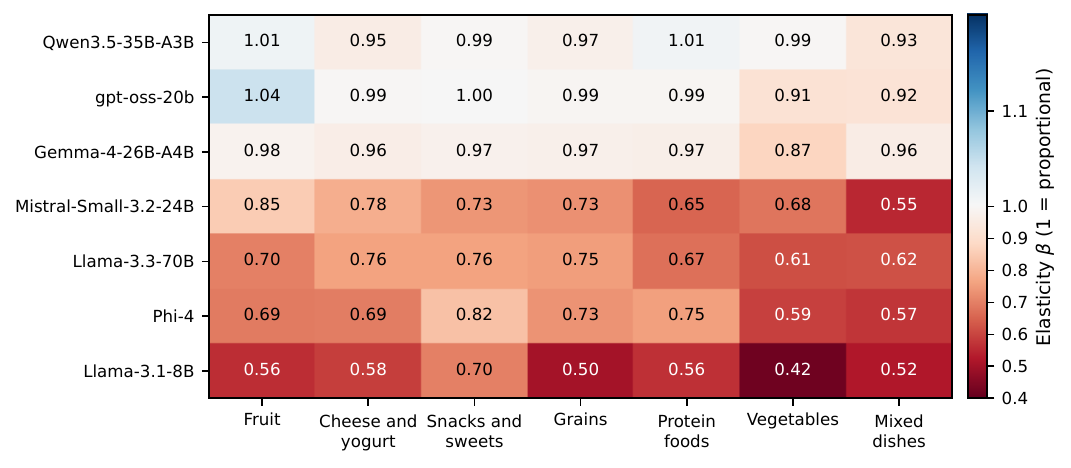}
\caption{Elasticity by food group and model, 24 foods per group and nine stated weights per food. Each cell is the elasticity $\beta$ estimated within the group; 1 is proportional scaling; cells below 1 are shaded red and cells above 1 blue. Models are ordered by overall elasticity. The comparison of groups is exploratory.}
\alttext{Heat map with seven food groups as columns and seven models as rows. The three near-proportional models have cells close to 1; the four severely compressing models have red cells throughout, darkest for mixed dishes and vegetables.}
\label{fig:groups}
\end{figure}

Llama-3.3-70B compressed less than Llama-3.1-8B: its elasticity was higher by 0.146 (95\% CI 0.106--0.185), and at 0.694 it still compressed severely. The mean curves of the severely compressing models fall steadily across the nine weights (Figure~\ref{fig:dose}); a positive quadratic term in three of them means that overestimation of small portions exceeds underestimation of large ones on the log scale. The curves differed across food groups in six models. In the exploratory comparison of groups, the most compressed group was mixed dishes or vegetables in every model (Figure~\ref{fig:groups}); in the four severely compressing models, these two groups had elasticities of 0.42 to 0.68. Without a stated amount, every model chose a larger serving than the FNDDS default, by a median factor of 1.31 to 1.67. Supplementary File 1 (section S5, Table S7) gives the curve-shape and default-serving results in full.

\subsection*{Robustness}

Compression persisted at full precision. With unquantized weights, Llama-3.1-8B, Phi-4 and Mistral-Small-3.2-24B compressed severely, with elasticities of 0.468 (95\% CI 0.433--0.507), 0.706 (0.677--0.735) and 0.757 (0.725--0.790), so compression replicated in all three. Against the 4-bit runs, the elasticity changed by $-0.081$, 0.014 and 0.048 (Supplementary File 1, Table S10): quantization moves the elasticity by up to 0.08, in a direction that differs between models, and leaves the severe compression of all three in place.

A plain question without a system prompt or answer format strengthened compression in five of the seven models (Supplementary File 1, Table S11). Six of seven models compressed severely, with elasticities of 0.294 to 0.910, and the criterion held in all four severely compressing models. Phi-4 fell from 0.692 to 0.294 and Llama-3.1-8B from 0.548 to 0.305, and Qwen3.5-35B-A3B, which scaled almost proportionally with the food-logging prompt, fell from 0.979 to 0.696. Only gpt-oss-20b, which reasons before it answers, scaled almost proportionally under both prompts (0.983 under the plain question).

With step-by-step reasoning, the four severely compressing models stated an energy per 100 g or per gram in 95.9\% to 98.2\% of answers. Their three-weight elasticities rose from 0.289 to 0.732 for the same question answered directly to 0.921 to 1.041: the CIs of Llama-3.3-70B and Phi-4 include 1, Mistral-Small-3.2-24B still compressed mildly, and Llama-3.1-8B remained severely compressing (Supplementary File 1, Table S12). In an exploratory comparison, reasoning brought the mean absolute log error at $0.25q$ and $4q$ to 0.23 to 0.63, against 0.40 to 0.91 for the direct greedy answers to the food-logging prompt; the two-step estimate reached 0.18 to 0.32.

\subsection*{Token probabilities}

The reader captured a median of 94.3\% to 100.0\% of the probability per prompt, and the elasticity of the distribution median lay within 0.05 of the sampled elasticity in all seven models, with differences of $-0.008$ to 0.044, so the probability distributions carry the compression (Table~\ref{tab:why}). The comparison used the 597 to 630 extreme-weight prompts per model whose $C$ and $R$ differed by at least 50\%. The compromise family fitted better than the switching family in every model: in the four severely compressing models, the mean difference in expected log-likelihood per food was 0.52 to 2.37, every 95\% CI lay above 0, and the result held without the 50\% filter and with other kernel bandwidths (Supplementary File 1, Tables S16 and S18). In a post hoc positive control built from each model's own references, synthetic probability that split between the two references within each prompt gave differences of $-0.28$ to $-1.18$, every one favoring switching, and synthetic blends gave 0.61 to 2.76; displacing the computed distribution by the observed distances in random directions also favored the compromise (0.42 to 1.40) (Supplementary File 1, Table S14). The comparison therefore shows that the probability does not split within a prompt, and the position of the probability, not the comparison, places the single value.

\begin{figure}[t]
\centering
\includegraphics[width=\textwidth]{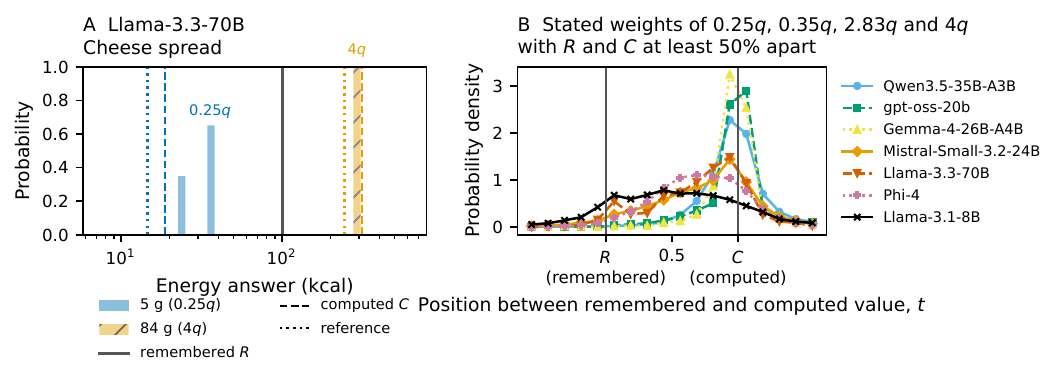}
\caption{Where the probability of an energy answer lies. (A) Probability distribution of the energy answer of Llama-3.3-70B for one food, cheese spread (American or Cheddar cheese base), at a quarter (blue) and four times (orange, hatched) the default amount, with the remembered value $R$ (median energy of the model's answer without an amount), the computed values $C$ (the median of the model's density answer times the weight; dashed) and the reference energies (dotted). The food is the one with the median likelihood difference for this model. (B) Mean probability density by position $t$ between the remembered value ($t = 0$) and the computed value ($t = 1$), over the prompts at $0.25q$, $0.35q$, $2.83q$ and $4q$ whose remembered and computed values differ by at least 50\%; line styles and markers identify the models.}
\alttext{Panel A is a histogram on a logarithmic energy axis: the probability at a quarter of the default amount sits on values above the computed and reference energies, and at four times the default amount on values between the reference energy and the computed value, with the remembered value between the two groups. Panel B shows line plots of probability density against position between the remembered and computed values: three models peak sharply at the computed value, three peak between the two values, closer to the computed value, and Llama-3.1-8B has a broad hump near the middle.}
\label{fig:why}
\end{figure}

In the four severely compressing models, the median of the distribution lay between the remembered and the computed values in 74\% to 81\% of prompts and beyond $C$ in 12\% to 21\% (Figure~\ref{fig:why}; Supplementary File 1, Table S15). Few of their prompts split the probability: 0.6\% to 7.1\% held at least 15\% of the probability both near $C$ and near $R$ or the model's energy per 100 g. The fitted weight on the computed value was 0.44 for Llama-3.1-8B and 0.71 to 0.80 for the other three, in the same order as their elasticities, against 0.98 to 1.00 in the near-proportional models. Part of the pull went to the model's energy per 100 g (Supplementary File 1, Table S17). The distributions signaled the error poorly: their central 80\% intervals contained the reference energy in 4\% to 67\% of prompts, and the width of the interval separated errors above 25\% from smaller ones with AUROCs of 0.52 to 0.71.

\begin{table}[t]
\centering
\caption{Token probabilities of the energy answers, 168 foods. Median: elasticity of the median of each answer's probability distribution; Sampled: elasticity of the sampled answers (Table~\ref{tab:main}). Difference: expected log-likelihood of the compromise account minus that of the switching account at $0.25q$, $0.35q$, $2.83q$ and $4q$, on the prompts whose computed value $C$ and remembered value $R$ differ by at least 50\% (597 to 630 per model; 166 foods for Gemma-4-26B-A4B and 167 for Llama-3.1-8B, whose other foods had no remembered value), mean per food with 95\% CI (food bootstrap); positive values favor the compromise. Simulated switching: the same difference when the probability of each prompt is replaced by a mixture of the model's own computed and remembered distributions at the compromise weight fitted to that prompt (Supplementary File 1, Table S14). Between: share of prompts whose distribution median lies between $R$ and $C$. Weight: median fitted weight of the compromise account on the computed value.}
\label{tab:why}
\small
\begin{singlespace}
\setlength{\tabcolsep}{4pt}
\begin{tabular}{@{}lcccccc@{}}
\toprule
& \multicolumn{2}{c}{Elasticity} & & Simulated & & \\
\cmidrule(lr){2-3}
Model & Median & Sampled & Difference (95\% CI) & switching & Between & Weight \\
\midrule
Qwen3.5-35B-A3B & 0.971 & 0.979 & 0.42 (0.33 to 0.52) & $-0.30$ & 57\% & 0.98 \\
gpt-oss-20b & 0.991 & 0.978 & 1.00 (0.75 to 1.25) & $-0.30$ & 51\% & 1.00 \\
Gemma-4-26B-A4B & 0.951 & 0.954 & 0.78 (0.58 to 0.98) & $-0.28$ & 59\% & 0.99 \\
Mistral-Small-3.2-24B & 0.753 & 0.709 & 0.95 (0.80 to 1.10) & $-0.85$ & 77\% & 0.80 \\
Llama-3.3-70B & 0.707 & 0.694 & 2.37 (2.10 to 2.64) & $-1.00$ & 74\% & 0.76 \\
Phi-4 & 0.709 & 0.692 & 1.88 (1.69 to 2.07) & $-1.18$ & 81\% & 0.71 \\
Llama-3.1-8B & 0.563 & 0.548 & 0.52 (0.42 to 0.63) & $-0.92$ & 80\% & 0.44 \\
\bottomrule
\end{tabular}
\end{singlespace}
\end{table}

\section*{DISCUSSION}

\subsection*{Principal findings}

When the weight of a food is stated exactly, the energy estimates of open-weight LLMs grow more slowly than the weight. All seven models compressed, and four compressed severely, with elasticities of 0.55 to 0.71, while staying close to the reference energy at the default amount itself. These four models stated energy densities with median errors of 11\% to 21\%, and three of them scaled almost proportionally once the density appeared in the prompt. A two-step estimate that multiplies the model's own density answer by the stated weight in code removed the dependence of the error on the weight. In the token probabilities of the severely compressing models, the probability rarely split between the energy of the model's remembered serving and the value computed from its own density, its median lay between the two in 74\% to 81\% of prompts, and the width of the distributions signaled the error poorly. At the default amount, the estimates of the severely compressing models averaged 1.02 to 1.10 times the reference energy, so an evaluation at one typical amount per food shows little bias.

\subsection*{Comparison with prior work}

The signature matches the flat-slope syndrome of human dietary recall \cite{gersovitz1978}. Here the weight is given in the prompt and restated correctly by the models, so the compression arises in the step from weight to energy. Image-based studies of LLM nutrition estimation report the same direction of error and attribute it to judging weight from a photograph \cite{fridolfsson2025,ohara2025,nakagawa2026}. The present results show that compression survives when the weight is known exactly, so part of the portion-size bias measured in image studies can come from the energy calculation that follows the weight judgment.

The closest mechanism is the under-use of supplied scale by vision-language models answering metric questions about video, with median log-log response slopes of 0.69 to 0.94 \cite{tan2026equisd}. The elasticity is the dose-response analogue of that slope, and its values under the food-logging prompt, 0.55 to 0.98, cover the same range in a text-only health task. In both settings the models hold the relevant knowledge and fail to apply it: the LLMs here stated energy densities with median errors of 9\% to 21\%, and six of the seven applied a supplied density correctly. The settings differ in the fix. Asking a vision-language model for a scale-free quantity with the reference still in the prompt did not remove the dependence on the reference \cite{tan2026equisd}; the two-step estimate asks for the density in a separate prompt that contains no weight, and the multiplication happens outside the model.

A single intermediate value has the form of the precision-weighted average of an observation and a prior that describes regression to the mean when LLMs estimate magnitudes \cite{ma2025bayesbench}; here the prior combines the model's typical serving and its energy per 100 g. In an early NutriBench version, most errors of GPT-3.5 with retrieved context came from using a retrieved value per 100 g without converting it to the amount asked \cite{hua2024nutribenchv1}; here a pull toward the energy of 100 g appears without any retrieved context. Because the whole distribution shifts, sampling several answers or reading a different quantile leaves the compression in place. LLMs that compare entities along a numerical attribute also answer from surface cues while holding the knowledge that would answer correctly, and thinking before answering moved them toward it \cite{lehmann2026shortcut}. Energy estimation for a stated weight shows the same split between knowing and applying, and the same remedy.

\subsection*{Implications}

Database-backed food loggers already compute the energy of a matched food as density times weight. The results argue for the same design when an LLM replaces the database: the model supplies the density, and code does the multiplication. Code then guarantees proportional scaling, and the accuracy of the estimate becomes the accuracy of the density, which a food composition database supplies for listed foods and the model supplies in one short answer for the rest. With the model's own density, the two-step estimate was at least as accurate as the direct estimate at the default amount in six of seven models, with mean absolute log errors of 0.18 to 0.32 against 0.20 to 0.37, so its gain at small and large portions costs little at typical ones (Supplementary File 1, Table S9). Qwen3.5-35B-A3B shows the dependence on the density: it scaled almost proportionally, its density answers ran a median 9\% below the reference values, and its two-step estimate was less accurate than its direct estimate at every tested weight (0.247 against 0.205 to 0.241); gpt-oss-20b's two-step estimate was less accurate at $4q$ (0.236 against 0.221). Step-by-step reasoning reached near-proportional scaling in three of the four severely compressing models, at the cost of a long answer per entry. Entries without a weight still need the direct estimate.

The error is systematic in portion size, so its burden is unequal. A person who logs a small portion, for example to restrict energy intake, receives an overestimate, and a person who logs a large portion, such as an athlete, receives an underestimate. For evaluation, accuracy at the amounts people usually eat does not certify a model for weighed entries. A dose-response probe of a fixed set of foods at several stated weights measures scaling in one number and needs no ground truth beyond a composition database.

\subsection*{Limitations}

The foods come from one US composition database, and the reference energies are database values for those foods as described. The elasticity is unaffected by a constant error in a reference density, which shifts only the absolute errors of the direct and two-step estimates. The outcome was energy; carbohydrate and the other nutrients follow the same density-times-weight arithmetic and were not measured. Each prompt named a single food, so meals with several components add a decomposition step that this study did not test, and the most compressed food groups were mixed dishes and vegetables. All prompts were in English. The size of the compression depended on the answer format: five of the seven models compressed more under a plain question than under the food-logging prompt. The elasticities describe these two formats, other formats can give other values, and which part of the food-logging format protects the estimate was not tested. The full-precision arm covered three models, and the reasoning arm covered the four severely compressing models at three weights with greedy decoding. The probability analysis fixed the answer to a single item, and its enumeration captured a median of 94.3\% of the probability for Llama-3.1-8B and at least 99.3\% for the other models. The density answers may draw on USDA values seen in training, so the two-step estimate may be less accurate for foods absent from public composition tables. The study evaluated open-weight models only, which makes every result reproducible from pinned weights; hosted commercial models, which many food-logging tools call, were not tested. Hosted models, multi-component meals, other languages and databases, and image pipelines that pass an estimated weight to a language model are the next tests; the dose-response probe and the two-step estimate apply to each without change.

\section*{CONCLUSION}

Open-weight LLMs asked for the energy of an exactly stated weight of food compress their estimates toward a typical serving, in a pattern that matches the flat-slope syndrome of human dietary recall. The compression is severe in four of the seven models, which overestimate a quarter of a default portion by 47\% to 130\% and underestimate four default portions by 20\% to 34\%, although they state energy densities with median errors of 11\% to 21\% and three of them apply a supplied density correctly. Food-logging tools that accept weighed entries should compute energy as density times weight, and evaluations of LLM nutrition estimation should test fixed foods across a range of stated weights.

\section*{ACKNOWLEDGMENTS}

The authors used a generative AI tool (Claude Code, Anthropic) \cite{claudecode2026} to assist with software development, literature search and writing. The authors reviewed all outputs and take full responsibility for the content of the article.

\section*{AUTHOR CONTRIBUTIONS}

DR: conceptualization, methodology, software, validation, formal analysis, investigation, data curation, visualization, writing (original draft), writing (review and editing) and project administration. FM: supervision and writing (review and editing). IB: project administration, supervision and writing (review and editing).

\section*{FUNDING}

This research received no specific grant from any funding agency in the public, commercial or not-for-profit sectors.

\section*{CONFLICTS OF INTEREST}

IB is an owner of Geometry Lab L.L.C-FZ, which operates Macropass, a commercial food-logging app that uses large language models to estimate the energy of logged foods. DR collaborates with Geometry Lab L.L.C-FZ on a research project and has no financial relationship with the company. Macropass and Geometry Lab L.L.C-FZ did not fund the study and had no role in its design, conduct, analysis or reporting, and the study evaluated no commercial product. FM declares no conflicts of interest.

\section*{DATA AVAILABILITY}

The protocols with their SHA-256 freeze record, the generation and analysis code, prompts, food samples, model responses, token-probability distributions and analysis outputs of this study are in a GitHub repository \cite{riccio2026repo}, under the MIT License for code and the Creative Commons Attribution 4.0 license for data. The repository is private until the article is published; until then, editors, reviewers and other researchers can request access from the corresponding author. FNDDS is a public-domain database of the US Department of Agriculture, and the model weights are public at the revisions listed in Supplementary File 1.

\section*{SUPPLEMENTARY MATERIAL}

\noindent Supplementary File 1: Prompts, food sampling frame, model revisions, development stage and additional results.

\noindent Supplementary File 2: TRIPOD-LLM reporting checklist.

{\raggedright

}

\end{document}